\documentclass[10pt,twocolumn,letterpaper]{article}

\usepackage[accsupp]{axessibility}
\usepackage{iccv}
\usepackage{times}
\usepackage{epsfig}
\usepackage{graphicx}
\usepackage{amsmath}
\usepackage{amssymb}
\usepackage{amsthm}
\usepackage{booktabs}
\usepackage{bm}
\usepackage{caption}
\usepackage{color}
\usepackage{multirow}
\usepackage[ruled,vlined]{algorithm2e}
\usepackage{url}
\usepackage[numbers,sort,compress]{natbib}

\newcommand{\myparagraph}[1]{\noindent\textbf{#1}}

\def\ie{\textit{i.e. }}
\def\eg{\textit{e.g. }}
\def\etc{\textit{etc. }}

\definecolor{Gray}{gray}{0.9}
\definecolor{green}{rgb}{0.55, 0.71, 0.0}
\definecolor{amaranth}{rgb}{0.9, 0.17, 0.31}
\definecolor{amber}{rgb}{1.0, 0.49, 0.0}
\definecolor{azure}{rgb}{0.0, 0.5, 1.0}
\definecolor{byzantine}{rgb}{0.74, 0.2, 0.64}
\definecolor{forestGreen}{rgb}{0.0, 0.54, 0.0}
\definecolor{blue}{rgb}{0.43, 0.71, 0.88}
\definecolor{pink}{rgb}{0.85, 0.44, 0.58}

\usepackage[pagebackref=true,breaklinks=true,letterpaper=true,colorlinks,bookmarks=false]{hyperref}

\usepackage[breaklinks=true,bookmarks=false]{hyperref}

\iccvfinalcopy % *** Uncomment this line for the final submission

\def\iccvPaperID{****} % *** Enter the ICCV Paper ID here

\ificcvfinal\fi

\begin{document}

%%%%%%%%% TITLE
\title{Probabilistic Human Mesh Recovery in 3D Scenes from Egocentric Views}

\newcommand*{\affaddr}[1]{#1} 
\newcommand*{\affmark}[1][*]{\textsuperscript{#1}}
\newcommand*{\email}[1]{\small{\texttt{#1}}}

\author{
Siwei Zhang\affmark[1] \quad
Qianli Ma\affmark[1,3] \quad
Yan Zhang\affmark[1] \quad
Sadegh Aliakbarian\affmark[2]  \quad
Darren Cosker\affmark[2]  \quad
Siyu Tang\affmark[1]\\
\affaddr{\affmark[1]ETH Z\"urich} \quad \affaddr{\affmark[2]Microsoft} \quad \affaddr{\affmark[3]Max Planck Institute for Intelligent Systems} \\
\email{\{siwei.zhang, qianli.ma, yan.zhang, siyu.tang\}@inf.ethz.ch} \\ \email{\{coskerdarren, saliakbarian\}@microsoft.com}
}

\twocolumn[{%
\renewcommand\twocolumn[1][]{#1}%
\maketitle
\begin{center}
    \newcommand{\teaserwidth}{\textwidth}
\vspace{-0.35in}
    \centerline{\includegraphics[width=0.95\linewidth]{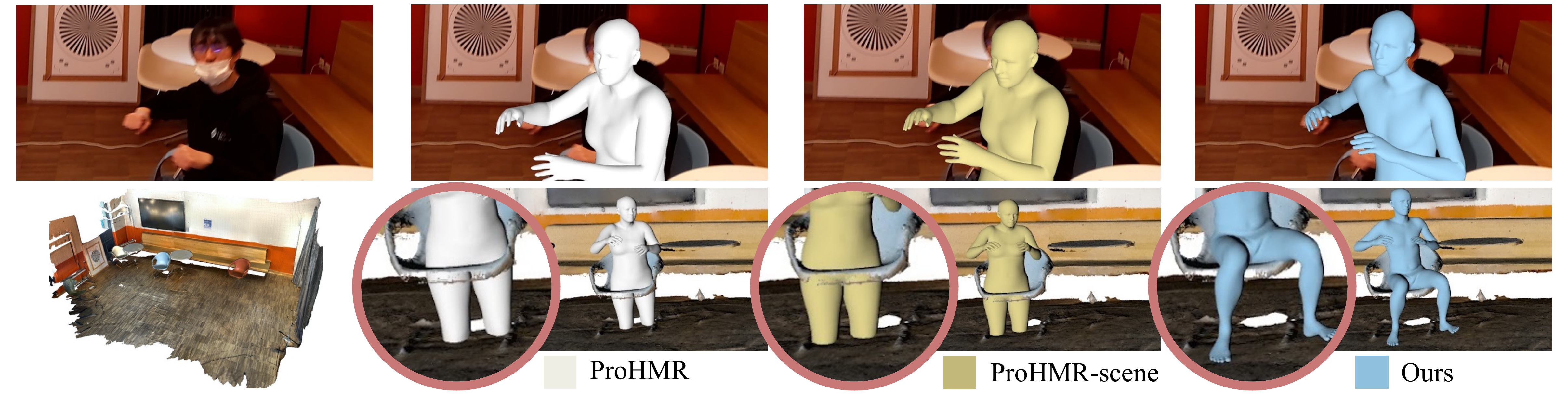}}
  % \vspace{-0.1in}
  \captionof{figure}{We propose EgoHMR, a novel scene-conditioned probabilistic method to recover the human mesh from an egocentric view image (typically with the body truncated) given the 3D environment. EgoHMR efficiently leverages the scene geometry and achieves significantly more plausible human-scene interactions compared to baseline methods, ProHMR~\cite{kolotouros2021probabilistic} and the extended ProHMR-scene (Sec.~\ref{sec:experiment-baselines}), even with severe body truncations.}
\label{fig:teaser}
% \vspace{-0.1in}
\end{center}%
}]

\maketitle
\ificcvfinal\thispagestyle{empty}\fi

%%%%%%%%% ABSTRACT
\begin{abstract}
% \vspace{-0.3cm}
Automatic perception of human behaviors during social interactions is crucial for AR/VR applications, and an essential component is estimation of plausible 3D human pose and shape of our social partners from the egocentric view. One of the biggest challenges of this task is severe body truncation due to close social distances in egocentric scenarios, which brings large pose ambiguities for unseen body parts. 
To tackle this challenge, we propose a novel scene-conditioned diffusion method to model the body pose distribution. 
Conditioned on the 3D scene geometry, the diffusion model generates bodies in plausible human-scene interactions, with the sampling guided by a physics-based collision score to further resolve human-scene inter-penetrations.
The classifier-free training enables flexible sampling with different conditions and enhanced diversity.
A visibility-aware graph convolution model guided by per-joint visibility serves as the diffusion denoiser to incorporate inter-joint dependencies and per-body-part control.
Extensive evaluations show that our method generates bodies in plausible interactions with 3D scenes, 
achieving both superior accuracy for visible joints and diversity for invisible body parts.
% %
The code is available at \url{https://sanweiliti.github.io/egohmr/egohmr.html}.

\end{abstract}

%%%%%%%%% BODY TEXT
\section{Introduction}
\label{sec:intro}

With the rapid development of Augmented and Virtual Reality (AR/VR) devices, understanding human actions, behaviors and interactions of our social partners (``\textit{interactee}") from the egocentric view is a vital component for head-mounted devices (HMDs) to truly become a virtual companion for humans. The first step towards the automatic perception of human behaviors and interactions for HMDs is to estimate 3D human pose and shape (\ie human mesh recovery) of the interactee from egocentic view images.

The research community has extensively studied human mesh recovery (HMR) from a single RGB image (usually captured with third-person view cameras) by predicting SMPL~\cite{loper2015smpl,pavlakos2019expressive} parameters from global image features \cite{kanazawa2018end, kolotouros2019cmr,Moon_2020_ECCV_I2L-MeshNet, Choi_2020_ECCV_Pose2Mesh, Kocabas_PARE_2021}.
However, their performance degrade significantly in egocentric images~\cite{zhang2022egobody}. 
A major challenge presented in egocentric scenarios is frequent body truncation~\cite{zhang2022egobody, liu20214d}, when individuals interact with each other within close proximity, while the HMD camera has a limited field-of-view. 
Since human poses are highly reliant on the surrounding environment, 3D scene structures can potentially provide strong cues to infer invisible body parts, which is crucial for precise understanding of human behaviors: sitting on a chair and standing on the ground may indicate distinct intentions and future behaviors.
Following PROX \cite{hassan2019resolving}, we assume a rough 3D scene structure is available, as such information nowadays can be easily obtained with commodity sensors. 
Furthermore, visual localization and mapping, one of the extensively studied topics in computer vision, is progressing rapidly for head-mounted devices \cite{sarlin2022lamar}. 
Therefore, we make the assumption that a coarse 3D model of the scene and the localization of the egocentric camera are readily available (\eg in HoloLens2~\cite{hololens2}).
In this paper, we focus on the challenging problem of estimating 3D human bodies that are heavily truncated from egocentric views due to the proximity between people and the motion of embodied cameras.

Given an egocentric image and the corresponding 3D environment, what should an ideal human mesh recovery method achieve?
In contrast to previous studies that mostly pursue full-body pose accuracy, we argue the following properties are desired in the egocentric scenario: 
(1) natural and plausible human-scene interactions;
(2) accurate body pose estimations consistent with image observations for visible joints;
(3) a comprehensive conditional distribution to generate {\it diverse} and {\it plausible} poses for unobserved body parts.
Several recent studies have attempted to tackle pose ambiguities caused by occlusions and body truncations. 
However, some methods can only produce a discrete number of body poses~\cite{li2019generating, oikarinen2021graphmdn, biggs20203d}, ignoring the continuous nature of the pose manifold. 
Other approaches model the continuous body pose distribution via conditional variational autoencoder~\cite{sharma2019monocular} or normalizing flows~\cite{kolotouros2021probabilistic, wehrbein2021probabilistic}, but with limited expressiveness.
Furthermore, the 3D environment is often ignored although it provides strong cues for inferring missing body parts in the image.
Existing works for human pose estimation~\cite{zhang2021learning, shimada2022hulc, shen2023learning} or generation~\cite{zhang2020place, hassan2021stochastic, zhao2023synthesizing, zhang2020generating} in 3D scenes typically cannot deal with truncated bodies in images.

To address these issues, we introduce a novel scene-conditioned probabilistic approach, the first method to recover human mesh in 3D scenes from the egocentric view image. 
Inspired by the recent diffusion models that can generate high fidelity images~\cite{dhariwal2021diffusion, song2020denoising, ho2020denoising, ho2022classifierfree} and human motions~\cite{ma2022mofusion, zhang2022motiondiffuse,kim2023flame,chen2023executing,tevet2023human} with flexible conditioning, our model is trained with a \textbf{conditional diffusion framework}, leveraging both the classifier-free guidance~\cite{ho2022classifierfree} and the classifier-guided diffusion sampling~\cite{sohl2015deep, song2021scorebased, dhariwal2021diffusion} for efficient scene conditioning.  
By training the model conditioning on a human-centric scene encoding, the diffusion denoiser can generate body poses with plausible human-scene interactions even with highly truncated bodies (see Fig.~\ref{fig:teaser}). 
The classifier-free training enables flexible sampling from models with or without image conditions, achieving accurate estimations of visible body parts while generating diverse plausible results for unseen body parts.
On top of that, the diffusion sampling process is guided by the gradients of a \textbf{physics-based collision score guidance} 
in the classifier-guided sampling manner, which further resolves human-scene inter-penetrations without requiring additional time-consuming postprocessing.

To facilitate the learning of the multimodal distributions of the invisible body parts, we introduce a \textbf{visibility-aware graph convolution model} as the diffusion denoiser network, to explicitly guide the network to learn more expressive distribution for invisible body parts.
Unlike existing methods~\cite{kanazawa2018end, li2022cliff, kolotouros2021probabilistic, wehrbein2021probabilistic} that simply condition the full body pose on a global image feature, we condition the pose diffusion of each body joint on the joint visibility mask and the 3D scene feature, in addition to the image feature. 
With such explicit visibility guidance, the model learns to 
estimate accurate body pose for visible joints while encouraging diversity for truncated body parts. 
We adopt a graph convolution network (GCN)~\cite{kipf2017gcn, zou2021modulated, zhang2020learning} to better incorporate local dependencies between highly-relevant body parts (\eg knee-foot) according to the human kinematic tree.

In summary, our contributions are: 
1) a novel scene-conditioned diffusion model for probabilistic human mesh recovery in the 3D environment from egocentric images;
2) a physics-based collision score that guides the diffusion sampling process to further resolve human-scene inter-penetrations;
3) a visibility-aware GCN architecture that incorporates inter-joint dependencies for the pose diffusion, and enables per-body-part control via the per-joint visibility conditioning.
% %
With extensive evaluations, the proposed method demonstrates superior accuracy and diversity of generated human bodies from egocentric images, in natural and plausible interactions with the 3D environment.

\section{Related Work}
\label{sec:related}

\myparagraph{Human mesh recovery from a single image.}
Given a single RGB image, the task of deterministic 3D human mesh recovery has been widely studied in the literature, with regression-based methods~\cite{tan2017indirect, kanazawa2018end, kolotouros2019cmr, omran2018neural, xu2019denserac, zhang2021body, zhou2021monocular, li2021hybrik, lin2021end-to-end, Moon_2020_ECCV_I2L-MeshNet, Choi_2020_ECCV_Pose2Mesh, Kocabas_PARE_2021, Kocabas_SPEC_2021, cho2022cross, li2022cliff, kolotouros2019convolutional, georgakis2020hierarchical}, optimization-based methods~\cite{pavlakos2019expressive, fang2021reconstructing, weng2021holistic, Bogo:ECCV:2016, lassner2017unite} or hybrid methods~\cite{kolotouros2019learning, song2020lgd, joo2020eft}, mostly adopting parametric 3D body models~\cite{Bogo:ECCV:2016, xu2020ghum} to represent the 3D human mesh.
Most of recent regression-based methods train neural networks to regress SMPL~\cite{Bogo:ECCV:2016} parameters from images~\cite{kanazawa2018end, Kocabas_SPEC_2021, li2022cliff, georgakis2020hierarchical, pavlakos2019texturepose}.
Model-free methods~\cite{lin2021end-to-end, Moon_2020_ECCV_I2L-MeshNet, Choi_2020_ECCV_Pose2Mesh, cho2022cross} regress human mesh vertices without relying on parametric body models. 
Optimization-based methods iteratively optimize the body model parameters to fit 2D keypoints, human silhouettes, \etc 
Some resort to self-contact~\cite{muller2021self}, scene~\cite{hassan2019resolving, weng2021holistic} or object~\cite{taheri2020grab} constraints to model interactions.
Hybrid methods seek to combine the strengths of both regression and optimization, for instance, SPIN~\cite{kolotouros2019learning}, which integrates an optimization loop into the deep learning framework. 
However, these methods often lack robustness when dealing with truncated bodies, a typical challenge in egocentric scenarios,
and cannot model the diverse nature of plausible missing body parts.
3D scene information is mostly disregarded, resulting in unrealistic human-scene interactions. 
In this work, we address these challenges by proposing a scene-conditioned generative model for human mesh recovery in 3D scenes.

\myparagraph{Multi-hypothesis for 3D human pose estimation.} 
Due to the limited image observation and depth ambiguity, estimating 3D human pose from a single image can have multiple potential solutions, especially when body truncation is presented. Recent works seek to model this task as a generative process or predict multiple hypothesis of possible poses.
A discrete number of hypotheses are generated in~\cite{jahangiri2017generating, li2019generating, oikarinen2021graphmdn, biggs20203d}.
To learn the continuous distribution of possible body poses,
conditional variational autoencoders~\cite{sharma2019monocular} or conditional normalizing flows~\cite{wehrbein2021probabilistic, kolotouros2021probabilistic} are adopted to generate unlimited number of hypotheses, but relying on additional scoring functions for pose selection~\cite{sharma2019monocular}, or discriminators to stabilize the training~\cite{wehrbein2021probabilistic, kolotouros2021probabilistic}.  
Nevertheless, the 3D scene constraint is usually not taken into account in this line of works.
We leverage the diffusion process to model the inherent pose ambiguity, with flexible scene-conditioning strategies.

\myparagraph{Egocentric human pose estimation.}
Most of existing works in egocentric human pose estimation targets the camera wearer's 3D pose~\cite{luo2020kinematics,jiang2017seeing,yuan2019ego,shiratori2011motion,tome2020selfpose,tome2019xr,guzov2021human}.
3D pose estimation for a social partner from the egocentric view has been receiving growing attention~\cite{ng2020you2me, liu20214d, ye2023decoupling}. However, the 3D scene geometry is neglected in~\cite{ng2020you2me, ye2023decoupling}, while~\cite{liu20214d} employs the time-consuming optimization, without modelling the probabilistic distribution of unseen body parts.

\myparagraph{Diffusion models for human motion/pose generation.}
Denoising Diffusion Models \cite{ho2020denoising,sohl2015deep,song2020denoising,song2021scorebased} are becoming a popular choice for human motion generative models.
Typically trained with the DDPM formulation~\cite{ho2020denoising} and based on Transformer-backbones, different methods vary on their choice of motion representation (e.g.~skeletons~\cite{zhang2022motiondiffuse,ma2022mofusion}, joint rotations~\cite{kim2023flame,tseng2023edge,alexanderson2023listen} or latent variables~\cite{chen2023executing}) and downstream tasks. 
To generate plausible human-scene interactions, SceneDiffuser~\cite{huang2023diffusion} condition the motion/pose diffusion model on 3D scenes and further guide the model with physics-based objectives at inference time.
In a similar fashion, PhysDiff~\cite{yuan2022physdiff} incorporates a physics-based motion projection module in the inference loop to resolve typical flaws in motion synthesis such as foot-ground penetrations. 
While these methods all target human action \textit{synthesis}, we take a different perspective and leverage Diffusion Models for a \textit{perception} task --- as a multi-proposal formulation to address the ambiguity in the egocentric pose estimation.

\section{Preliminaries}
\myparagraph{SMPL body model~\cite{loper2015smpl}.}  
We model the human mesh with the parametric SMPL body model.
SMPL parametrizes a body as a function $\mathcal{M}_b(\boldsymbol{\gamma}, \boldsymbol{\beta}, \boldsymbol{\theta})$ of the global translation $\boldsymbol{\gamma} \in \mathbb{R}^3$, body shape $\boldsymbol{\beta} \in \mathbb{R}^{10}$, and full body pose $\boldsymbol{\theta}\in \mathbb{R}^{24\times 3}$ of 23 body joints, plus the global orientation, returning a mesh $\mathcal{M}_b$ with 6890 vertices.

\myparagraph{Conditional Diffusion Model.}
\label{sec:diffusion_model_general}
We use DDPM~\cite{ho2020denoising} as our formulation of pose diffusion model. 
DDPM learns a distribution of body poses $\boldsymbol{\theta}$ through a forward diffusion process and an inverse denoising (sampling) process. 
The forward diffusion process is a Markov chain of added Gaussian noise over $t\in\{1,\cdots,T\}$ steps:
\begin{equation} \label{eq:ddpm_forward}
    q(\boldsymbol{\theta}_t|\boldsymbol{\theta}_{t-1}) = \mathcal{N}(\boldsymbol{\theta}_t; \sqrt{\alpha_t}\boldsymbol{\theta}_{t-1}, (1-\alpha_t)\mathbf{I}),
\end{equation}
where the variance $1-\alpha_t\in(0,1]$ increases with $t$ according to a pre-defined schedule, and $T$ is the total number of noise corruption steps.

At the core of the inverse process is a denoiser neural network $D(\cdot)$, which is trained to remove the added Gaussian noise based on 
condition signal $c$ at each step $t$. 
% \begin{equation}\label{eq:ddpm_reverse}
% p_\theta(x_{t-1}|x_t) = \mathcal{N}()
% \end{equation}
Following the design of recent motion diffusion models~\cite{tevet2023human}, the denoiser predicts the clean signal itself: 
$\hat{\boldsymbol{\theta}}_0 = D(\boldsymbol{\theta}_t, t, \boldsymbol{c})$.

The Gaussian forward diffusion process has a special property that allows directly sampling $\boldsymbol{\theta}_t$ from $\boldsymbol{\theta}_0$: 
\begin{equation}\label{eq:ddpm_forward2}
% q(x_t|x_0) = \mathcal{N}(x_t; \sqrt{1-\beta_t}x_{t-1}, \beta_t\mathbf{I}),
\boldsymbol{\theta}_t = \sqrt{\alpha_t}\boldsymbol{\theta}_0 + \sqrt{1-\alpha_t}\boldsymbol{\epsilon},~\textrm{where}~\boldsymbol{\epsilon}\sim\mathcal{N}(0, \mathbf{I}).
\end{equation}
With this, training the denoiser amounts to sampling a random $t\in \{0,\cdots, T-1\}$, adding noise according to Eq.~\eqref{eq:ddpm_forward2}, and optimizing the \textit{simple objective}~\cite{ho2020denoising}:
\begin{equation}\label{eq:loss_simple}
\mathcal{L}_\textrm{simple} = \mathit{E}_{\boldsymbol{\theta}_0\sim q(\boldsymbol{\theta}_0|\boldsymbol{c}), t\sim [1, T]}\left[ \|\boldsymbol{\theta}_0 - D(\boldsymbol{\theta}_t, t, \boldsymbol{c})\|_2^2 \right] .
\end{equation}

In this paper we use the subscript $t$ to denote the diffusion timestep and superscript $j$ for the body joint index.

\section{Method}
\label{sec:method}

\begin{figure*}
\centering
\includegraphics[width=\textwidth]{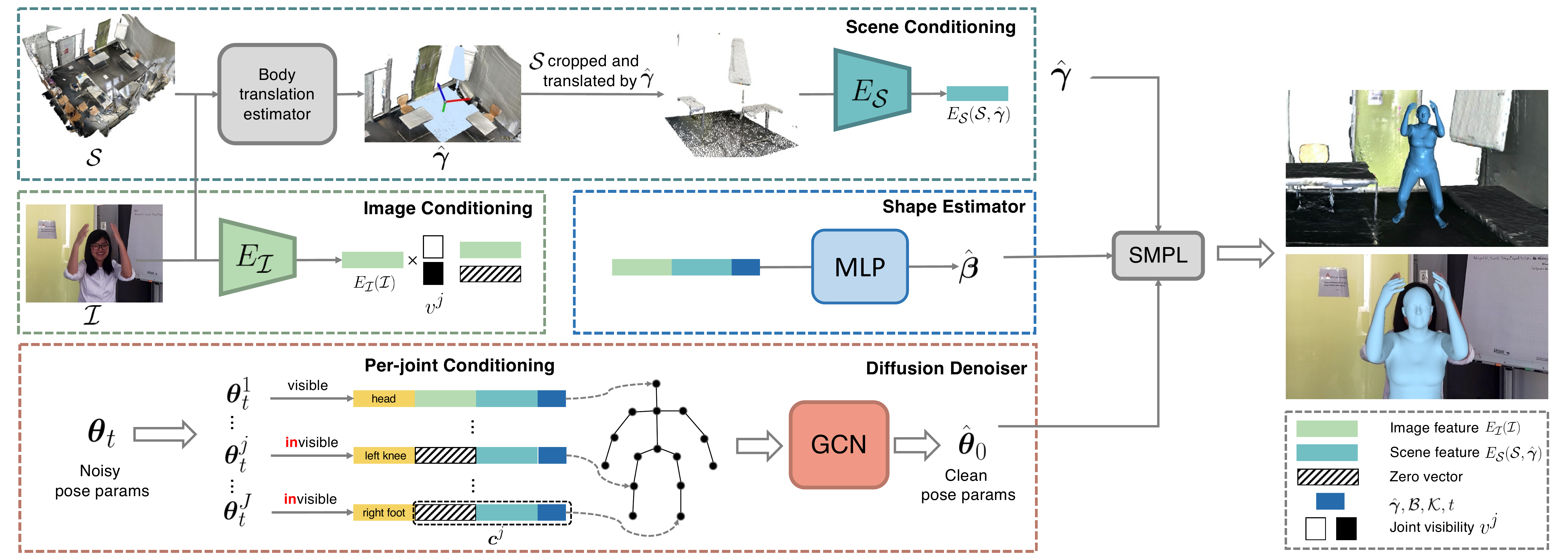}
\caption{\textbf{Method overview.} Given an egocentric image $\mathcal{I}$ containing a partially visible body and the 3D scene geometry $\mathcal{S}$, the proposed model generates a SMPL mesh with plausible human-scene interactions and conforming to the image observation. 
The model comprises a Scene Conditioning module to estimate the body translation $\hat{\boldsymbol{\gamma}}$ and encode the local scene features, an Image Conditioning module to encode image features, a GCN-based Diffusion Denoiser to predict the clean body pose $\hat{\boldsymbol{\theta}}_0$ from noisy pose $\boldsymbol{\theta}_t$ with Per-joint Conditioning, and a Shape Estimator to estimate body shape $\hat{\boldsymbol{\beta}}$.
}
\label{fig:overview}
\end{figure*}

We introduce a novel scene-conditioned diffusion method to recover the human mesh in 3D environments from the egocentric image.
Fig.~\ref{fig:overview} shows the overview of our model. 
Given an egocentric image $\mathcal{I}$ with a truncated body, the camera intrinsics $\mathcal{K}=(f, c_x, c_y)$, and the 3D scene point cloud $\mathcal{S}\in \mathbb{R}^{N\times 3}$ with $N$ points, our goal is to learn the conditional distribution of body poses $p(\boldsymbol{\theta} | \mathcal{I}, S, \mathcal{K})$, with the objective of generating humans that naturally interact with the scene while being aligned with the image observation. The body translation $\boldsymbol{\gamma}$ and shape $\boldsymbol{\beta}$ are learnt in a deterministic way following~\cite{kolotouros2021probabilistic}.
Here $f$ and $(c_x, c_y)$ denote the focal length and the camera center, respectively. 

The diffusion denoising process (Sec.~\ref{sec:diffusion_model_general}) is modeled by a visibility-aware graph convolution network (GCN), which achieves per-body-part control with the per-joint conditioning on the input image, scene geometry and joint visibility (Sec.~\ref{sec:diffusion-conditions}).
The diffusion model is trained in a classifier-free manner~\cite{ho2022classifierfree}, enabling flexible sampling from both models that include or exclude image conditions to enhance diversity and expressiveness for truncated body parts.
Guided by the gradients of a physics-based human-scene collision score, we further resolve human-scene inter-penetrations during the diffusion sampling (Sec.~\ref{sec:diffusion-sampling}). The training objectives are explained in Sec.~\ref{sec:method_training_loss}.

\subsection{Scene-conditioned Pose Diffusion Model}
\label{sec:diffusion-conditions}

\myparagraph{3D scene conditioning.}
To model fine-grained human-scene interactions, only a small region of the scene surrounding the human is relevant. An accurate global body translation provides better reasoning about local scene geometries, thus crucial for accurate and scene-plausible local body pose modelling.
To this end, we propose a body translation estimator (see Fig.~\ref{fig:overview}) to estimate the translation $\hat{\boldsymbol{\gamma}}$ by extracting features from $\mathcal{I}$ and the full 3D scene point cloud $\mathcal{S}$. The global scene features alleviate the scale ambiguity and localize the body in the 3D environment accurately (please refer to Supp.~Mat.~for more details).
Given the estimated body translation $\hat{\boldsymbol{\gamma}}$, scene vertices in a $2\times2m$ region around the human are selected and considered as the scene conditioning for the diffusion denoising process. 
Furthermore, the selected scene vertices are translated by $\hat{\boldsymbol{\gamma}}$ such that the scene vertex origin is around the body pelvis. 
The translated scene point cloud is encoded by a scene encoder $E_{\mathcal{S}}$ into a localized human-centric scene feature $E_{\mathcal{S}}(\mathcal{S}, \hat{\boldsymbol{\gamma}})$ for modelling detailed interactions between the body and the 3D scene.

\myparagraph{Per-joint conditioning.}
A ResNet50 backbone~\cite{he2016deep} $E_\mathcal{I}$ is used to encode the cropped input image $I$ into a latent feature $E_\mathcal{I}(\mathcal{I}) \in \mathbb{R}^{2048}$ as the image conditioning. However, regressing the full body pose directly from the global image feature as in previous works~\cite{kanazawa2018end,kolotouros2019learning,kolotouros2021probabilistic} loses the explicit joint visibility information. 
With the body truncated in the image, different properties are desired for different body parts:
the model should achieve pose accuracy for observed body joints, and pose expressiveness for truncated body parts. 
To achieve the per-body-part control, the 2D joint visibility mask $\mathcal{V} = (v^0, ..., v^j, ..., v^{J-1}) \in \mathbb{R}^{J} $ is extracted from the image via OpenPose~\cite{openpose} 2D joint detections, and passed to the diffusion model as the visibility condition. Here $J$ denotes the joint number; and $v^j$ equals to 1 if the $j-$th joint is visible in $\mathcal{I}$, otherwise 0.

Apart from above conditions, the bounding box information $\mathcal{B}$ is additionally fed to the network inspired by~\cite{li2022cliff}, serving as a global-location-aware feature to better model global information:
\begin{equation}
    \mathcal{B} = (b_x, b_y, b) / f,
\end{equation}
where $b_x, b_y, b$ denote the bounding box center coordinates and bounding box size in the full image, respectively. 
The full condition $\boldsymbol c^j$ for $j$-th joint is formulated as the concatenation of: the image feature (masked out by $v^j$ if the joint is not visible), 3D scene feature, estimated body translation, bounding box feature, camera intrinsics and the diffusion timestep $t$:
\begin{equation}
    \boldsymbol c^j = (E_\mathcal{I}(\mathcal{I}) \cdot v^j, E_\mathcal{S}(\mathcal{S}, \hat{\boldsymbol{\gamma}}), \hat{\boldsymbol{\gamma}}, \mathcal{B}, \mathcal{K}, t).   
\end{equation}
By conditioning on a per-joint basis, the model achieves flexible control over each body part according to the joint visibility. This allows for both accurate pose estimation of visible body parts, and flexibility for invisible body parts.

\myparagraph{Diffusion denoiser architecture.}
At diffusion timestep $t$, for the noisy body pose parameters $\boldsymbol{\theta}_t$, the diffusion denoiser $D$ predicts the clean pose parameter $\hat{\boldsymbol{\theta}}_0$ conditioned on $\boldsymbol{c}$. 
Previous works either use a single MLP for the full body pose~\cite{kanazawa2018end, li2022cliff, kolotouros2021probabilistic, wehrbein2021probabilistic} with redundant dependencies between irrelevant joints, or employ a separate MLP for each joint~\cite{Kocabas_PARE_2021} which fully ignores joint dependencies.
Instead, we adopt a graph convolution network (GCN)~\cite{kipf2017gcn, zou2021modulated} as the denoising function $D$. Given the predefined human kinematic tree, GCN models inter-joint dependencies by treating the human skeleton as a graph, where each node corresponds to a body joint. 
Specifically, for each joint $j$, the noisy pose parameters $\boldsymbol{\theta}^j_t$ is embedded and concatenated with the per-joint conditioning $\boldsymbol{c}^j$ as the feature for node $j$ in the GCN (see Fig.~\ref{fig:overview}).
We adopt the modulated GCN~\cite{zou2021modulated} for the implementation. 
Different from vanilla GCN~\cite{kipf2017gcn}, the modulated GCN learns node-specific modulation vectors to disentangle joint feature transformations, such that different relations can be learnt between joints. 
It also modulates the affinity matrix with learnable parameters to model additional edges beyond the human skeleton~\cite{zou2021modulated}. Please refer to Supp.~Mat. for implementation details.

% shape prediction
\myparagraph{Shape estimator.} As the body shape $\boldsymbol{\beta}$ is a global feature and more robust to body truncations, an MLP branch (the body shape estimator in Fig.~\ref{fig:overview}) predicts $\hat{\boldsymbol{\beta}}$ from the conditions $(E_{\mathcal{I}}(\mathcal{I}), E_S(\mathcal{S}, \hat{\boldsymbol{\gamma}}), \hat{\boldsymbol{\gamma}}, \mathcal{B}, \mathcal{K})$ in a deterministic way.

\subsection{Scene-guided Sampling}
\label{sec:diffusion-sampling}
We use the DDPM~\cite{ho2020denoising} sampler at inference time. 
At each sampling step $t$, the denoiser predicts $\hat{\boldsymbol{\theta}}_0 = D(\boldsymbol{\theta}_t, t, \boldsymbol{c})$, which is noised back to $\boldsymbol{\theta}_{t-1}$ by sampling from the Gaussian distribution:
\begin{equation}
\label{eq:ddpm_reverse_sample}
    \boldsymbol{\theta}_{t-1} \sim \mathcal{N}(\mu_t (\boldsymbol{\theta}_t, \hat{\boldsymbol{\theta}}_0), \Sigma_t),
\end{equation}
where $\mu_t (\boldsymbol{\theta}_t, \hat{\boldsymbol{\theta}}_0)$ is a linear combination of $\boldsymbol{\theta}_t$ and $\hat{\boldsymbol{\theta}}_0$, and $\Sigma_t$ is a scheduled Gaussian distribution as in~\cite{ho2020denoising}. The sampling is iteratively repeated from $t=T-1$ until $t=0$.

\myparagraph{Classifier-free guidance.}
The diffusion model $D$ is trained with classifier-free guidance~\cite{ho2022classifierfree} by randomly masking out the image feature $E_\mathcal{I}(\mathcal{I})$ for all joints with a probability of 5\% during training, such that the model also learns the pose distribution independent from the image, enabling flexible sampling. 
By combining the poses of visible joints sampled from the full-conditioned model and the poses of invisible joints sampled from the model excluding image conditions:
\begin{equation}
\label{eq:ddpm_reverse_sample_fuse}
    \hat{\boldsymbol{\theta}}_0 = D(\boldsymbol{\theta}_t, t, \boldsymbol{c}) \odot \mathcal{V} + D(\boldsymbol{\theta}_t, t, \boldsymbol{c}_{\phi}) \odot (1-\mathcal{V}) ,
\end{equation}
we achieve better per-body-part control, with enhanced diversity for invisible body parts while maintaining the accuracy of visible body parts (see Sec.~\ref{sec:experiment-ablation} for the ablation study). Here $\odot$ denotes multiplying the joint visibility mask with the pose of each joint, and $\boldsymbol{c}_{\phi}$ is the condition excluding image features.

\myparagraph{Collision score guided sampling.}
The proposed scene encoding in Sec.~\ref{sec:diffusion-conditions} implicitly incorporates the 3D environment constraint into the diffusion conditioning. However, the network cannot fully resolve the human-scene collisions in a fine-grained way. 
Another way to incorporate conditions into the diffusion model is the score-function based guidance. As shown in previous works~\cite{sohl2015deep, song2021scorebased, dhariwal2021diffusion}, a pre-trained classifier can be adopted to provide class information by adding the gradients of the classifier at each sampling step to the diffusion model, to guide the sampling process towards the class label. 
Inspired by this, we propose a collision score guided sampling strategy to further resolve implausible human-scene inter-penetrations, by leveraging implicit body representations~\cite{mihajlovic2022coap, LEAP:CVPR:21}.
We use COAP~\cite{mihajlovic2022coap} to model neural articulated occupancy of human bodies as the zero-level set $f_{\Theta}(q|\mathcal{G})=0$, where $q$ is the input query point and $\mathcal{G(\boldsymbol{\theta}, \boldsymbol{\beta})}$ is the input bone transformations. By querying each scene vertex near the body to see if it is inside of the human volume, the collision score is defined as:
\begin{equation}
\label{eq:coap_loss}
    \mathcal{J}(\boldsymbol{\theta}) = \frac{1}{|\mathcal{S}|} \sum\nolimits_{q \in \mathcal{S}} \sigma(f_{\Theta}(q|\mathcal{G})) \mathbb{I}_{f_{\Theta}(s|\mathcal{G}) > 0},
\end{equation}
where $\sigma(\cdot)$ stands for the sigmoid function. The gradient $\nabla \mathcal{J}(\boldsymbol{\theta})$ efficiently guides the diffusion sampling process to further alleviate the human-scene collisions (see Sec.~\ref{sec:experiment-ablation} for the ablation study) by modifying Eq.~\eqref{eq:ddpm_reverse_sample} to:
\begin{equation}
\label{eq:ddpm_reverse_sample_coll_guidance}
    \boldsymbol{\theta}_{t-1} \sim \mathcal{N}(\mu_t (\boldsymbol{\theta}_t, \hat{\boldsymbol{\theta}}_0) + a \Sigma_t \nabla \mathcal{J}(\boldsymbol{\theta_t}), \Sigma_t),
\end{equation}
where the guidance is modulated by $\Sigma_t$ and a scale factor $a$.

\subsection{Training Objectives}
\label{sec:method_training_loss}
% \todo{maybe move to the end of method section}
The diffusion model is trained with $\mathcal{L}_\textrm{simple}$ in Eq.~\eqref{eq:loss_simple} for body pose $\hat{\boldsymbol{\theta}}_0$, together with the the 3D joint loss $\mathcal{L}_\textrm{3D}$, the 2D keypoint re-projection loss $\mathcal{L}_\textrm{2D}$ calculated in the full image frame, and the shape loss $\mathcal{L}_{\beta}$:
\begin{equation}
\begin{aligned}
    \mathcal{L}_\textrm{3D} &= \|FK(\hat{\boldsymbol{\theta}}_0, \hat{\boldsymbol{\beta}}) - FK(\boldsymbol{\theta}_0, \boldsymbol{\beta})\|^2 , \\
    \mathcal{L}_\textrm{2D} &= \|\Pi_\mathcal{K}(J_\textrm{3D} + \hat{\boldsymbol{\gamma}}) - J_\textrm{est}\|^2, \\
    \mathcal{L}_{\beta} &= \|\hat{\boldsymbol{\beta}} - \boldsymbol{\beta}\|^2 .
\end{aligned}
\end{equation}
Here $FK(\cdot)$ denotes the SMPL joint regressor to obtain 3D joint coordinates. 
$J_\textrm{est}$ is the 2D body keypoint detections, and $\Pi_\mathcal{K}$ denotes the 3D to 2D projection in the full image with camera intrinsics $\mathcal{K}$.
The overall loss is defined as:
\begin{equation}
\label{eq:training_objective}
\begin{aligned}
    \mathcal{L} &= \mathcal{L}_\textrm{simple} + \lambda_\textrm{3D} \mathcal{L}_\textrm{3D} + \lambda_\textrm{2D} \mathcal{L}_\textrm{2D} +  \lambda_{\boldsymbol{\beta}} \mathcal{L}_{\boldsymbol{\beta}}  \\
    &+ \lambda_\textrm{coll} \mathcal{L}_\textrm{coll} + \lambda_\textrm{orth} \mathcal{L}_\textrm{orth} ,
\end{aligned}
\end{equation}
where $\mathcal{L}_\textrm{orth}$ is a regularizer to force the generated 6D pose rotation representation to be orthonormal, $\mathcal{L}_\textrm{coll}$ is the scene collision loss the same as defined in Eq.~\eqref{eq:coap_loss}, and $\lambda$s are the corresponding weight factors for each loss term.

\section{Experiments}
\label{sec:experiment}

\subsection{Dataset}
Most of existing datasets for 3D human pose evaluation are either captured from the third-person view with the full body visible~\cite{ionescu2013human3, von2018recovering,  mono-3dhp2017, hassan2019resolving}, or do not contain the 3D environment~\cite{ng2020you2me}. 
The most relevant dataset is \textbf{EgoBody}~\cite{zhang2022egobody}, a large-scale egocentric dataset capturing 3D human motions during social interactions in 3D scenes. 
Each sequence involves two people interacting with each other in 3D environments, and one subject wears a head-mounted camera to capture the egocentric images containing the other \textit{interactee}. 
Due to the close social interaction distance, the captured egocentric images typically exhibit strong body truncations for the interactee.
We use the official train / test splits for training and evaluation, which include 90,120 / 62,140 frames, respectively.

\subsection{Evaluation Metrics}
% \vspace{-0.2cm}
Strong body truncations pose inherent ambiguity for invisible body parts, while the visible body parts are relatively deterministic given the image observations, thus evaluating the accuracy of full body joints together as in~\cite{kolotouros2021probabilistic, biggs20203d} is not suitable for such tasks. We propose to evaluate the proposed method from the following aspects.

\myparagraph{Accuracy.}
We employ Mean Per-Joint Position Error (MPJPE) and Vertex-to-Vertex (V2V) errors in \textit{mm} to evaluate the accuracy of estimated body pose and shape. For MPJPE, we report different variations: G-MPJPE as the joint error in the global coordinate; MPJPE as the pelvis-aligned joint errors, and PA-MPJPE as the joint errors after Procrustes Alignment~\cite{gower1975generalized}. 
We report mean MPJPE and V2V metrics over $n$ samples for visible joints. 
Nevertheless, such evaluation protocol assumes a single correct ground truth prediction, which does not hold for invisible joints. Instead of the mean error, we report \textit{min-of-n} MPJPE for invisible body parts following~\cite{biggs20203d}. To be specific, among the generated $n$ samples for each input image, we select the one with the minimum MPJPE for all evaluated methods.

\myparagraph{Physical plausibility.}
We evaluate the collision and contact scores between the body mesh and the 3D scene. 
As the clean scene signed distance field (SDF) is not available in EgoBody, we define the collision score as the ratio of the number of scene vertices inside of the body mesh to the number of all scene vertices. 
Note that the absolute scale of the collision score may vary with different size of the 3D scene, but within the same scene surface, the relative comparison between different methods indicates different collision levels. 
We use the cropped scene as described in Sec.~\ref{sec:diffusion-conditions} for evaluation (with a fixed size of 20,000 vertices). 
Following~\cite{huang2023diffusion}, the contact score is defined as 1 if the body contacts with the scene in a distance within a threshold (2 \textit{cm}), otherwise 0. 
We report the mean collision score and contact score over all test samples. 
Note that a model with better physical plausibility should achieve \textit{both} a lower collision score and a higher contact score.

\myparagraph{Diversity.}
Diversity is evaluated for invisible body parts to demonstrate the expressiveness of the methods, by calculating the standard deviation (std) and Average Pairwise Distance (APD) of invisible body joints~\cite{huang2023diffusion}. A higher diversity is desired, yet there is usually a \textit{trade-off} between diversity and accuracy/plausibility.
An ideal model should predict diverse poses for invisible body parts, with accurate visible body poses and plausible human-scene interactions.

\subsection{Baselines and Our Method}
\label{sec:experiment-baselines}
As there is no existing work for our task, we compare our method with the following baselines: 
(1) ProHMR~\cite{kolotouros2021probabilistic}, the state-of-the-art method for probabilistic human mesh recovery from a single image based on conditional normalizing flows;
(2) ProHMR-scene, in which we enhance the ProHMR framework by incorporating the scene geometry $\mathcal{S}$ with a global scene feature encoder.
The ProHMR framework is trained with 2D and 3D supervisions for the sample drawn from $z=0$ (referred as the `mode'), such that the `mode' can be used as a predictive model (see~\cite{kolotouros2021probabilistic} for details). Different training weights for the samples drawn from $z\neq0$ yields a trade-off between the accuracy and diversity for the generated multiple hypotheses. Here $z$ denotes the latent space of the normalizing flow. 
To demonstrate this trade-off, we train the ProHMR-scene baseline with different training configurations. For samples drawn from $z\neq0$, we supervise with:
1) only 2D re-projection loss (the same as in~\cite{kolotouros2021probabilistic}), denoted as \textit{-orig};
2) 3D supervision with small weights, denoted as \textit{-weak-3D};
3) 3D supervision with large weights, denoted as \textit{-strong-3D}. Here the 3D supervision refers to the loss on 3D body joints and SMPL parameters. 
Please refer to Supp.~Mat. for details of the baseline models.

\begin{table*}[tb]
\centering
\footnotesize
\caption{\textbf{Evaluation for accuracy, physical plausibility and diversity on EgoBody.} Here \textit{-vis} and \textit{-invis} denote visible and invisible joints, respectively. `coll' and `contact' refer to the collision and contact scores, respectively. 
The percentage for the collision score denotes how much improvement the corresponding method has achieved compared with the ProHMR baseline.
The best results are in boldface.}
\label{tab:results}
\scalebox{0.92}{
\begin{tabular}{clccccccccc}
\toprule[1pt]

\multirow{2}{*}{$n$} & \multirow{2}{*}{Method} 
& G-MPJPE $\downarrow$ & MPJPE $\downarrow$ & PA-MPJPE $\downarrow$ & V2V $\downarrow$ & \textit{min-of-n} MPJPE $\downarrow$ & \multirow{2}{*}{coll $\downarrow$}  & \multirow{2}{*}{contact $\uparrow$}  & std $\uparrow$ & APD $\uparrow$  \\
& & -\textit{vis} & -\textit{vis} & -\textit{vis} & -\textit{vis} & -\textit{invis}  &  &  & -\textit{invis} & -\textit{invis} \\

\midrule
\multirow{3}*{5} & ProHMR~\cite{kolotouros2021probabilistic}   & 181.33  & 72.92  & \textbf{46.57}  & 90.88 & 128.56 & 0.00346 (-) & 0.94 & 26.53 & 32.95\\
& ProHMR-scene-\textit{orig}   & 133.72  & 73.13  & 48.87  & 91.59 & 117.88 & 0.00305 (12\%) & 0.96 & \textbf{35.42} & \textbf{44.02} \\
& Ours    & \textbf{128.62}  & \textbf{65.70}  & 47.17  & \textbf{82.93} & \textbf{107.77} & \textbf{0.00191 (45\%)}  & \textbf{0.99} & 29.59 & 35.15 \\

\midrule
\multirow{3}*{10} & ProHMR~\cite{kolotouros2021probabilistic}   & 181.70  & 73.65  & \textbf{47.06}  & 91.84 & 121.67 & 0.00346 (-) & 0.94 & 29.26 & 35.06  \\
& ProHMR-scene-\textit{orig}   & 134.57  & 74.42  & 49.63  & 93.21 & 110.64 & 0.00305 (12\%) & 0.96 & \textbf{39.07} & \textbf{46.86} \\
& Ours    & \textbf{128.60}  & \textbf{65.69}  & 47.20  & \textbf{82.94} & \textbf{100.75} & \textbf{0.00191 (45\%)}  & \textbf{0.99} & 29.59 & 35.15 \\

\midrule
\multirow{5}*{20} & ProHMR~\cite{kolotouros2021probabilistic}   & 181.87  & 73.99  & 47.29  & 92.28 & 115.48 & 0.00346 (-) & 0.93 & 30.55 & 36.14 \\
& ProHMR-scene-\textit{orig}   & 134.97  & 75.03  & 49.98  & 93.98 & 103.81 & 0.00306 (12\%) & 0.96 & \textbf{40.82} & \textbf{48.32} \\
& ProHMR-scene-\textit{weak-3D}   & 131.51  & 67.81  & 46.94  & 85.05 & 111.68 & 0.00313 (10\%) & 0.96 & 21.79 & 25.67 \\
& ProHMR-scene-\textit{strong-3D}   & 134.40  & 66.42  & \textbf{45.96}  & 82.99 & 123.34 & 0.00316 (9\%) & 0.97 & 13.01 & 15.23 \\
& Ours    & \textbf{128.60}  & \textbf{65.69}  & 47.19  & \textbf{82.94} & \textbf{94.96} & \textbf{0.00191(45\%)} & \textbf{0.99} & 30.23 & 35.14 \\

 \bottomrule[1pt]
 \end{tabular}
 }
\end{table*}

\begin{figure*}[t]
\vspace{1cm}
    \centering
    \includegraphics[width=\linewidth]{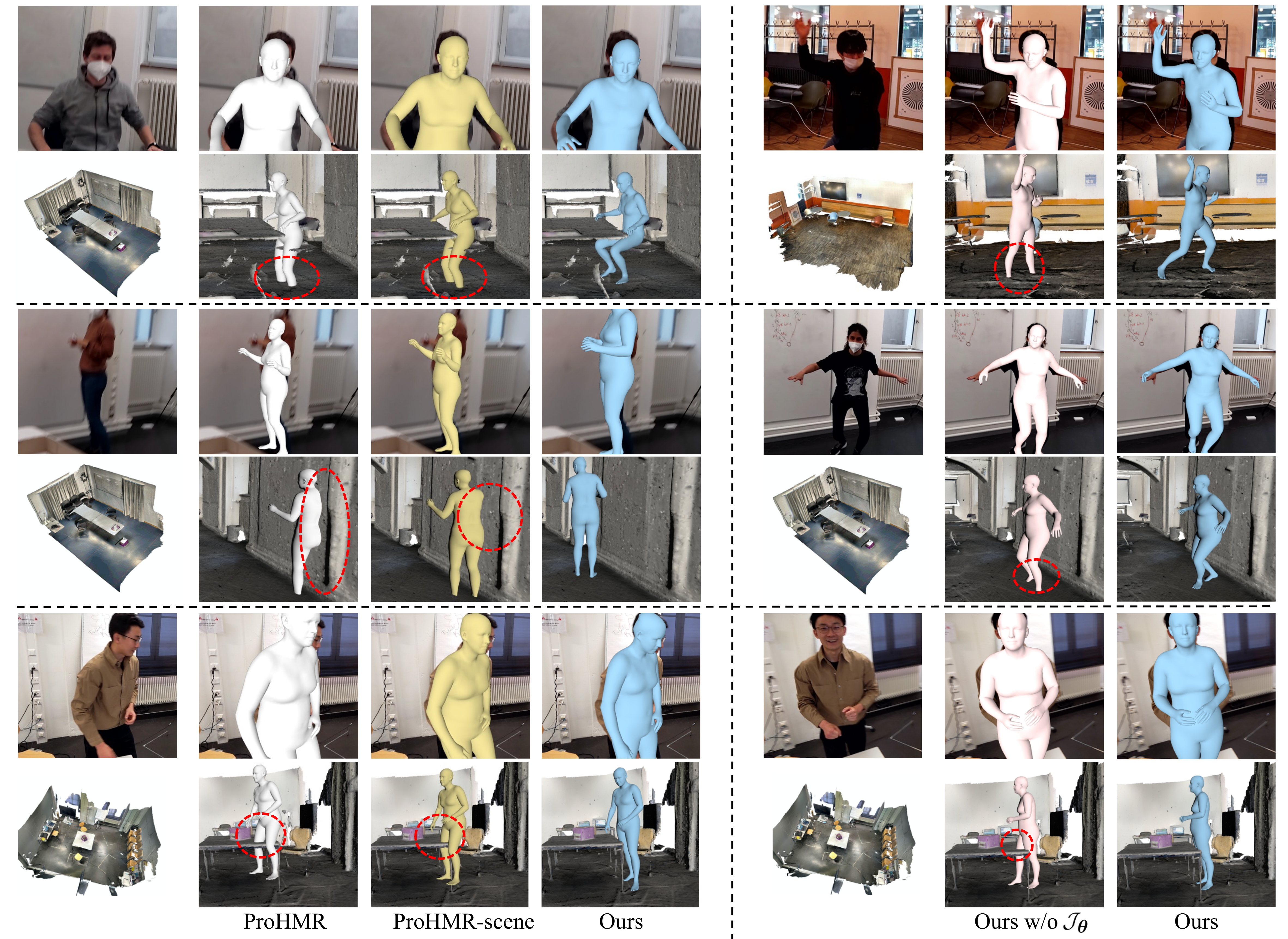}
    \caption{\textbf{Qualitative results on EgoBody dataset.} The left part shows the comparison between the baseline methods and our proposed method. The right part demonstrates the effectiveness of our proposed scene collision score guidance $\mathcal{J}_{\boldsymbol{\theta}}$ to further resolve human-scene inter-penetrations. \textcolor{red}{Red} circles point to human-scene collisions.}
    \label{fig:qualitative_examples}
\end{figure*}

\begin{table*}[tb]
\centering
\footnotesize
\caption{\textbf{Ablation study.} 
$\mathcal{J}_{\boldsymbol{\theta}}$ and CF denote the collision score guidance and classifier-free guidance during the diffusion sampling (Sec~\ref{sec:diffusion-sampling}). $\mathcal{L}_\textrm{coll}$ denotes the collision loss for training (Eq. (\ref{eq:training_objective})). All results are reported for $n=20$.}
% \vspace{-0.2cm}
\label{tab:ablation}
\begin{tabular}{lcccccc}
\toprule[1pt]

\multirow{2}{*}{Method} 
& MPJPE $\downarrow$  & \textit{min-of-n} MPJPE $\downarrow$ & \multirow{2}{*}{coll $\downarrow$}  & \multirow{2}{*}{contact $\uparrow$}  & std $\uparrow$ & APD $\uparrow$  \\
& -\textit{vis} & -\textit{invis}  &  &  & -\textit{invis} & -\textit{invis} \\

\midrule
Ours  & 65.69 & \textbf{94.96} & \textbf{0.00191} & \textbf{0.99} & \textbf{30.23} & \textbf{35.14} \\
Ours w/o $\mathcal{J}_{\boldsymbol{\theta}}$  & 65.09  & 97.20  & 0.00225  & 0.99 & 21.53 & 25.32 \\
Ours w/o $\mathcal{J}_{\boldsymbol{\theta}}$ w/o CF & \textbf{64.58}  & 98.82  & 0.00229  & 0.99 & 15.75 & 19.56 \\
Ours w/o $\mathcal{J}_{\boldsymbol{\theta}}$ w/o CF w/o $\mathcal{L}_\textrm{coll}$ & 64.91  & 96.26  & 0.00302  & 0.99 & 22.94 & 21.79 \\

 \bottomrule[1pt]
 \end{tabular}
\end{table*}

\begin{figure*}[t]
    \centering
    \includegraphics[width=0.9\linewidth]{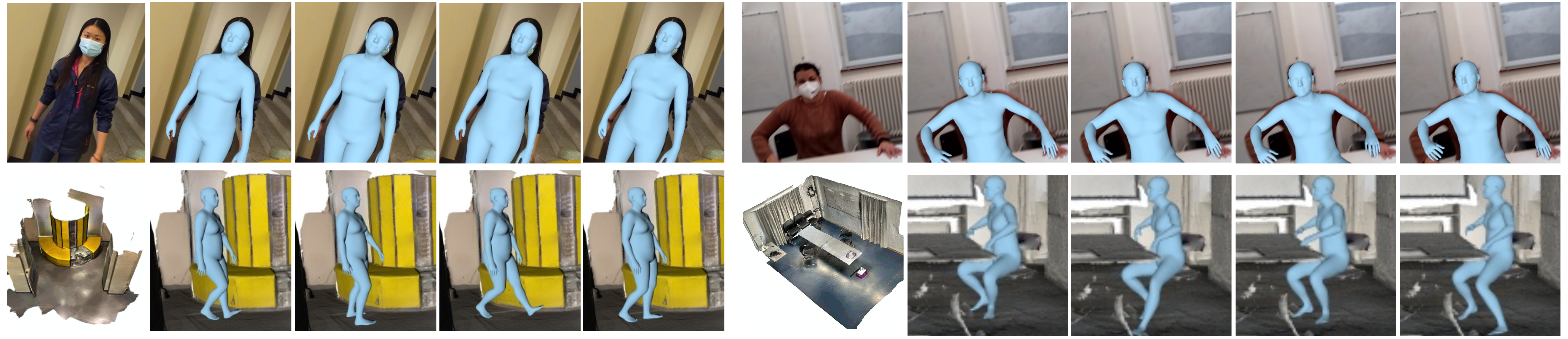}
    \caption{\textbf{Multiple samples.} Given the same input, our method can generate diverse poses for invisible body parts.}
    \label{fig:diversity_examples}
    \vspace{0.2cm}
\end{figure*}

\subsection{Results}
Tab.~\ref{tab:results} shows the quantitative results. 
Compared to the baselines, our method achieves the best G-MPJPE, MPJPE and V2V metrics for visible joints, demonstrating accurate estimations of the global translation, local body pose for visible joints and body shape.
For truncated body joints, our method obtains the best \textit{min-of-n} MPJPE, indicating that our generated results better covers the ground truth distribution for truncated body parts than baselines, thanks to the efficient scene conditioning.
Our method also outperforms the baselines with the higher contact ratios and significantly better collision scores. In particular, our collision score decreased \textbf{45\%} compared with ProHMR, and \textbf{37\%} compared with the best ProHMR-scene model. This indicates that the proposed method can effectively leverage the scene conditions and resolve body-scene inter-penetrations, while achieving better contact relations between the body and the 3D environment. 
Fig.~\ref{fig:qualitative_examples} (left) shows the qualitative results and the comparison with the baseline method.

From the results among different ProHMR-scene baselines ($n=20$) in Tab.~\ref{tab:results}, we can clearly see a trade-off between the accuracy and diversity: stronger 3D supervision (ProHMR-scene-\textit{strong-3D}) has better accuracy for visible body parts (MPJPE-\textit{vis}), yet lower diversity (std-\textit{invis} and APD-\textit{invis}) for invisible joints. 
The reason behind this is that the ProHMR framework, like many other existing works, condition the generative process of full body pose only on the global image feature. 
In contrast, our method leverages the joint visibility and the GCN architecture, achieving the best MPJPE for visible joints, diverse poses for invisible joints, and natural human-scene interactions (see Fig.~\ref{fig:diversity_examples}), while the baseline with a similar diversity level (ProHMR) and the baseline with the highest diversity (ProHMR-scene-\textit{orig}) exhibit much higher MPJPE errors and more severe human-scene collisions.
There is a very slight increase for PA-MPJPE of our model compared to the ProHMR-scene-\textit{strong}/\textit{weak-3D} baselines, while we obtain a much higher diversity for invisible body joints.
This reveals that our method achieves flexible per-body-part control to balance between the accuracy and diversity for different body parts.

To validate the reconstruction accuracy, the full-body accuracy of the proposed method is also evaluated and compared against existing deterministic baselines on EgoBody dataset.
The results show that our method even achieves higher full-body accuracy than deterministic methods, which demonstrates that the proposed model achieves diversity without sacrificing the reconstruction accuracy.
In addition, the proposed method is also evaluated on PROX~\cite{hassan2019resolving} dataset, a monocular RGB-based third-person view dataset with human-scene interactions, and our method outperforms the basline methods in terms of pose accuracy, human-scene interaction plausibility and pose diversity. 
Please refer to Supp.~Mat. for detailed results of evaluations on PROX and comparison with deterministic methods on EgoBody.

% \vspace{-0.15cm}
\subsection{Ablation Study}
% \vspace{-0.1cm}
\label{sec:experiment-ablation}
To investigate the influences of different training and sampling schemes for the diffusion model, we perform an ablation study and the results are shown in Tab.~\ref{tab:ablation}.
The scene collision score efficiently resolves the human-scene inter-penetrations from two different perspectives: via the collision loss $\mathcal{L}_\textrm{coll}$ during training and via the explicit collision score guidance $\mathcal{J}_{\boldsymbol{\theta}}$ during the diffusion sampling (Eq.~(\ref{eq:ddpm_reverse_sample_coll_guidance})).
Fig.~\ref{fig:qualitative_examples} (right) shows that the collision score guidance $\mathcal{J}_{\boldsymbol{\theta}}$ improves the human-scene interaction relstionships.
The classifier-free guidance during the diffusion sampling in Eq.~(\ref{eq:ddpm_reverse_sample_fuse}) significantly improves the sample diversity, by leveraging the invisible joints sampled from the model trained excluding the image conditions. 
It is worth mentioning that the collision score guidance $\mathcal{J}_{\boldsymbol{\theta}}$ also contributes to the sample diversity. 
This is not surprising as the gradients of $\mathcal{J}_{\boldsymbol{\theta}}$ can potentially optimize the body pose to different directions given different initial body poses.

Compared with the trade-off in the ProHMR-based methods where the accuracy sacrifices a lot for diversity (and vice versa), the classifier-free guidance and the collision score guidance in our model only bring marginal changes for the MPJPE of visible body parts, in the trade-off for significantly enhanced physical plausibility, sample diversity and better \textit{min-of-n} MPJPE for unobserved joints.

\vspace{-0.2cm}
\section{Conclusion}
\vspace{-0.1cm}
\label{sec:conclusion}
This paper introduces a novel scene-conditioned probabilistic approach for human mesh recovery from the egocentric image given the 3D scene geometry. 
The proposed diffusion model efficiently leverages the scene conditions and joint visibility to achieve per-body-part control and model human-scene interactions. 
The classifier-free guidance and the collision score guidance enables flexible sampling process with further enhanced diversity and physical plausibility. 
We demonstrate that our model generates human bodies in realistic interactions with the 3D environment, with diverse poses for invisible body parts while conforming to the image observation.
Nevertheless, the current method also has limitations. For example, currently only a single frame is considered as the input. A promising future direction is to further exploit temporal information to reconstruct temporally consistent human motions from the egocentric view.

{
\myparagraph{Acknowledgements.} 
This work was supported by Microsoft Mixed Reality \& AI Zurich Lab PhD scholarship.
Qianli Ma is partially funded by the Max Planck ETH Center for Learning Systems. 
We sincerely thank Korrawe Karunratanakul, Marko Mihajlovic and Shaofei Wang for the fruitful discussions.
}

% \clearpage
{\small
\bibliographystyle{ieee_fullname}
\bibliography{egbib}
}

\begingroup
\onecolumn 
\clearpage

\appendix

\begin{center}
\Large{\bf Probabilistic Human Mesh Recovery in 3D Scenes from Egocentric Views \\
% \vspace{0.3cm} 
**Supplementary Material**\\
}
\vspace{0.2cm}
\vspace{1cm}
\end{center}

\setcounter{page}{1}
\setcounter{table}{0}
\setcounter{figure}{0}
\renewcommand{\thetable}{S\arabic{table}}
\renewcommand\thefigure{S\arabic{figure}}

\section{Architecture Details}
The detailed model architecture is illustrated in Fig.~\ref{fig:appendix-architecture}. 
The local scene encoder $E_{\mathcal{S}}$ is an MLP network consisting of several residual blocks to encode the cropped input scene point cloud of $M$ points (translated by the estimated body translation $\hat{\boldsymbol{\gamma}}$ from the camera coordinate system) into a 512-d scene feature. 
In the diffusion denoiser $D$, for each joint $j$, we use the 6D representation~\cite{zhou2019rotation} to represent the joint rotations. A linear layer first maps the input noisy pose parameters $\boldsymbol{\theta}^j_t$ into a 512-d pose embedding . The timestep $t$ is embedded by an MLP with the sinusoidal function. 
The pose embedding is concatenated with the corresponding context embedding (including the image feature, scene feature, timestep embedding, $\mathcal{B}$, $\mathcal{K}$, and the estimated body translation $\hat{\boldsymbol{\gamma}}$) as the input feature for node $j$ in the GCN.
The GCN module consists of an input GCN layer, followed by four residual modulated GCN blocks~\cite{zou2021modulated} and a final GCN layer, which outputs the clean pose parameters $\hat{\boldsymbol{\theta}}^j_0$ for each joint $j$.

\begin{figure*}[h]
\centering
\includegraphics[width=0.9\textwidth]{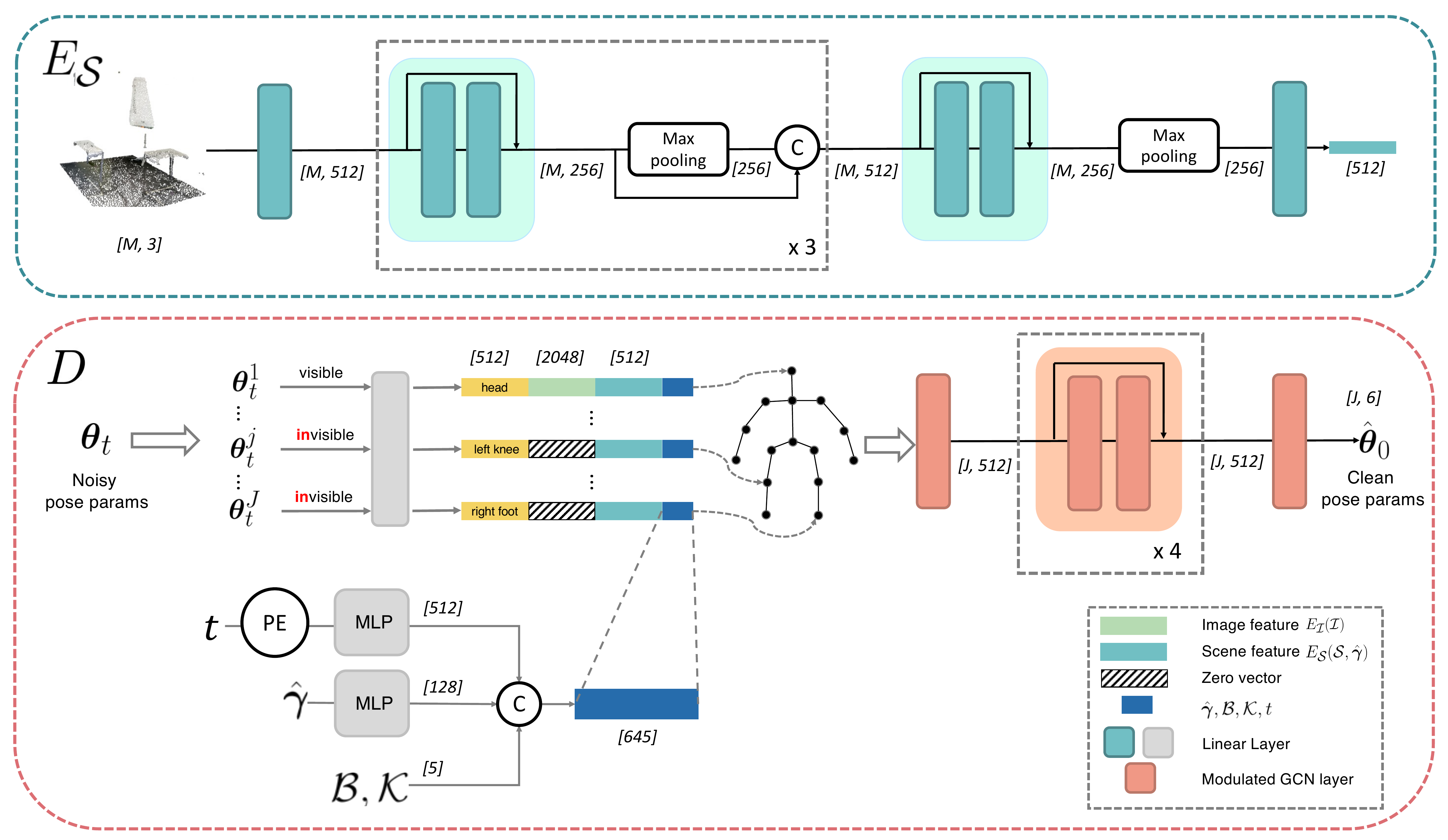}
\caption
{
\textbf{Architecture details for the local scene encoder $\boldsymbol{E_{\mathcal{S}}}$ and the diffusion denoise $\boldsymbol{D}$}. Numbers in `[]' indicates the corresponding feature dimension. PE stands for positional encoding.
}
\label{fig:appendix-architecture}
\end{figure*}

\section{Body Translation Estimator}
As the body translation is coupled with the body shape and pose parameters in the SMPL model, an accurate estimation of $\boldsymbol{\gamma}$ also relies heavily on the learning of the body pose $\boldsymbol{\theta}$ and shape $\boldsymbol{\beta}$. We adopt ProHMR~\cite{kolotouros2021probabilistic} as the backbone for the body translation estimator, which estimates ($\boldsymbol{\gamma}$, $\boldsymbol{\theta}$, $\boldsymbol{\beta}$) jointly, but with three major modifications.

First, the scale ambiguity between $\boldsymbol{\gamma}$ and $\boldsymbol{\beta}$ poses great challenges to the accurate global translation prediction in~\cite{kolotouros2021probabilistic} with a single image as the input. 
We leverage the 3D scene point cloud $\mathcal{S}$ with a global scene encoder to extract the global scene feature. The global scene encoder has the same architecture as the local scene encoder in Fig.~\ref{fig:appendix-architecture}, but with the full scene point cloud $\mathcal{S}$ in the camera coordinate system as the input. The encoded global scene feature is concatenated with the image feature (encoded by a ResNet50 backbone~\cite{he2016deep}) as the conditioning input to the normalizing flow.
Second, existing works~\cite{kolotouros2021probabilistic, kanazawa2018end, Kocabas_PARE_2021} take the cropped image (containing the target person, resized to a fixed resolution) as the input, which discards the location information in the full image camera coordinate system. The ignorance of the original camera introduces additional ambiguity and results in inaccurate estimations of global information. 
Besides the cropped image features, we additionally feed the bounding box information $\mathcal{B}$ (the same as in Eq.~(4) in the main paper) to the network to provide global-aware features.
On top of that, the predicted body is transformed back to the original camera coordinate system and the 2D keypoint reprojection loss is calculated in the full image instead of the cropped image. The projected 2D keypoints in the original image have similar perspective distortions with the person in the full image, offering better supervision for 3D predictions~\cite{li2022cliff}.
Last but not least, the model is further conditioned on the camera intrinsics $\mathcal{K}$ such that it can be adapted to different cameras and headsets. 

Here the supervision on body pose $\boldsymbol{\theta}$ and body shape $\boldsymbol{\beta}$ provides auxiliary information for the accurate estimation of the body translation. 
And we employ this straightforward scene-conditioned model as the baseline method ProHMR-scene in Sec.~5.3 in the main paper.
However, the 3D scene features here are not sufficiently localized to learn fine-grained human-scene interactions for local body pose (as demonstrated in Sec.~5.4 in the main paper), thus we only take the predicted translation $\hat{\boldsymbol{\gamma}}$ from this model, and propose the scene-conditioned pose diffusion model for better local pose reasoning.

To train the body translation estimator we employ the same training objectives as in~\cite{kolotouros2021probabilistic}, but with the 2D keypoint re-projection loss calculated in the full image frame. The trained model also serves as the ProHMR-scene-\textit{orig} baseline in Sec.~5.3 in the main paper.

\section{Implementation Details}
\myparagraph{Training details.}
The image encoder $E_{\mathcal{I}}(\mathcal{I})$ is loaded from the pretrained weights from~\cite{kolotouros2021probabilistic} (for our method and all baseline methods). 
For the ProHMR baseline, we load the entire pretrained checkpoint from~\cite{kolotouros2021probabilistic} and fine-tune it on EgoBody dataset.
In our model, the body translation estimator and the local pose diffusion model are trained separately. The diffusion model is trained with the ground truth body translation $\boldsymbol{\gamma}$. During inference, we use the predicted $\hat{\boldsymbol{\gamma}}$ from the body translation estimator to crop and translate the local scene point cloud to encode the local scene feature, and feed $\hat{\boldsymbol{\gamma}}$ as the input to the diffusion model.
The body pose $\boldsymbol{\theta}$ is transformed from the 6D representation to the rotation matrix to calculate $\mathcal{L}_\textrm{simple}$.
The weights for $\mathcal{L}_\textrm{simple}, \mathcal{L}_\textrm{3D}, \mathcal{L}_\textrm{2D}, \mathcal{L}_{\boldsymbol{\beta}}, \mathcal{L}_\textrm{coll}, \mathcal{L}_\textrm{orth}$ are 0.001, 0.05, 0.01, 0.0005, 0.0002, and 0.1, respectively. The collision loss term $\mathcal{L}_\textrm{coll}$ is disabled for the first three epochs.
The model is trained with a single TITAN RTX GPU of 24GB memory for approximately 18 epochs, with a batch size of 12, which takes around 24 hours.

\myparagraph{Collusion score guided sampling.}
In Eq.~(9) in the main paper, we set $a$ as 2. For the last 10 diffusion denoising timesteps, we ignore the $\Sigma_t$ and only scale $\nabla \mathcal{J}(\boldsymbol{\theta_t}), \Sigma_t)$ by $a$ such that the collision score guidance does not diminish too much at the end of the sampling process.

\myparagraph{Evaluation protocol.}
For the evaluation, the standard PA-alignment is obtained from the full body joints, however in the highly truncated case the diverse nature of invisible body parts could deviate from the ground truth and result in inaccurate PA-alignment, thus we perform PA-alignment with only visible body joints and report the PA-MPJPE metric.

\section{More Experiments}
\subsection{Ablation Study on Model Architecture and Per-joint Conditioning}
We also conduct experiments with the following two architectures as the diffusion denoiser $D$ to verify the effectiveness of our proposed per-joint visibility conditioning strategy and the GCN architecture:
1) a single MLP network to predict the full body pose conditioned on $ \boldsymbol c = (E_\mathcal{I}(\mathcal{I}), E_\mathcal{S}(\mathcal{S}, \hat{\boldsymbol{\gamma}}), \hat{\boldsymbol{\gamma}}, \mathcal{B}, \mathcal{K}, t)$, \ie the image feature, the scene feature, the bounding box information, the camera intrinsics and the diffusion timestep, without the per-joint visibility mask, denoted as `full-body MLP'; 
2) using the same per-joint conditioning strategy as our proposed method, but with $J$ MLP networks to predict the pose parameters for each body joint separately, where the MLPs share the same architecture but with different weights, denoted as `per-joint MLP'. 
The results are shown in Tab.~\ref{fig:appendix-architecture}. 

For the per-joint MLP model, there is a significant drop on sample diversity for invisible body parts, as such model architecture disables the classifier-free guidance. 
With the classifier-free guidance, the pose for invisible body parts sampled from the model excluding the image condition can be fused into the standard sampling results, thus improving the sample diversity for invisible body parts.
With a separate MLP for each body joint independently from other joints, each MLP for the invisible joint already excludes the image condition, therefore no additional classifier-free guidance can be applied on top of that to further improve diversity. 
Different from the GCN architecture which considers the human kinematic tree, the per-joint MLP model neglects the inter-joint dependencies, which are crucial to model human poses and human-scene interactions. Due to this reason, the \textit{min-of-n} MPJPE for invisible joints of the per-joint MLP model is also higher compared to the proposed GCN architecture, indicating that the generated body pose for visible body parts cannot cover the ground truth distribution well enough. 

For the full-body MLP model, the full body pose is conditioned on the image and scene feature. Without the explicit per-joint visibility information, the network can hardly achieve the precise per-body-part control, therefore struggling to balance between the accuracy and diversity for different body parts (with the lowest diversity compared with other two models in Tab.~\ref{tab:appendix-ablation}). On the contrary, our proposed per-joint conditioning strategy can leverage the joint visibility to achieve both accuracy for visible joints and diversity for invisible joints, together with plausible human-scene interactions.

\begin{table*}[t]
\centering
% \footnotesize
\caption{\textbf{Ablation study for model design choices.} All experiments are conducted without the scene collision score guidance $\mathcal{J}(\boldsymbol{\theta_t})$. The results are reported for $n=5$.}
\label{tab:appendix-ablation}
\begin{tabular}{lcccccc}
\toprule[1pt]

\multirow{2}{*}{Method} 
& MPJPE $\downarrow$  & \textit{min-of-n} MPJPE $\downarrow$ & \multirow{2}{*}{coll $\downarrow$}  & \multirow{2}{*}{contact $\uparrow$}  & std $\uparrow$ & APD $\uparrow$  \\
& -\textit{vis} & -\textit{invis}  &  &  & -\textit{invis} & -\textit{invis} \\

\midrule
Ours  & 65.10 & \textbf{107.59} & \textbf{0.00225} & \textbf{0.99} & \textbf{20.30} & \textbf{25.34} \\
Per-joint MLP &  65.35 & 113.62 & \textbf{0.00225} & \textbf{0.99} & 16.50 & 20.51 \\ 
Full-body MLP &  65.65 & 111.67 & 0.00228 & 0.98 & 11.08 & 12.98 \\

 \bottomrule[1pt]
 \end{tabular}
\end{table*}

\subsection{Evaluation on PROX Dataset}
We also evaluate the trained model on PROX dataset, a third-person view dataset with monocular RGB frames for human-scene interaction scenarios.
Due to the large domain gap of camera-body distances between EgoBody (1\(\sim \)3\textit{m}) and PROX (2\(\sim \)5\textit{m}), predicted body translations are not accurate on PROX for all methods since they are trained on EgoBody.
To better analyze our model's capability on scene-conditioned 3D body estimation on PROX, we report the numbers with ground truth body translations for all methods. 
Our model outperforms the baselines (Tab.~\ref{tab:eval_prox}), with more accurate local pose, more plausible interactions with the scene and relatively high diversity for unseen body parts.

\begin{table}[t]
\centering
\caption{
\textbf{Evaluation on PROX with trained models on EgoBody.} 
All results are reported for $n=5$. 
}
 % \vspace{-0.2cm}
\label{tab:eval_prox}
\begin{tabular}{lcccccc} 
\toprule[1pt]
% \hline

\multirow{2}{*}{Method} 
& MPJPE $\downarrow$  & \textit{min-of-n} MPJPE $\downarrow$ & coll $\downarrow$  & std $\uparrow$ & APD $\uparrow$  \\
& -\textit{vis} & -\textit{invis}  &  & -\textit{invis} & -\textit{invis} \\

\midrule
ProHMR-scene-\textit{orig}  & 117.97  & 217.69  & 0.00907  & \textbf{48.88} & \textbf{59.49} \\
ProHMR-scene-\textit{weak-3D} & 115.53  & 201.37  & \underline{0.00839}  & 25.69 & 31.68 \\
ProHMR-scene-\textit{strong-3D} & \underline{112.37}  & \underline{199.33}  & 0.00887   & 20.63 & 25.26 \\
Ours  & \textbf{107.17} & \textbf{198.72} & \textbf{0.00739} & \underline{30.01} & \underline{37.35} \\

% \hline
 \bottomrule[1pt]
 \end{tabular}
 % \vspace{-0.2cm}
\end{table}

\subsection{Comparison with Deterministic Methods}
Here we show the full-body accuracy of deterministic baselines (Tab.~\ref{tab:eval_deterministic}): they lag behind our method by a considerable margin. 
Results show that even conditioning the network with scene features (SPIN-scene, by encoding scene point clouds with an additional scene feature on top of SPIN) cannot perform comparably with our method.
This validates (1) the advantage of our probabilistic formulation with highly ambiguous poses; 
(2) feeding the network with scene features alone is insufficient, 
and our scene-guided diffusion sampling effectively addresses this.

\begin{table}[t]
\centering
% \scriptsize
\caption{\textbf{Comparison with Deterministic baseline.} Here the MPJPE and PA-MPJPE are calculated for the full body.}
 % \vspace{-0.2cm}
\label{tab:eval_deterministic}
\begin{tabular}{cccccc}
% \hline
\toprule[1pt]
Metrics & METRO~\cite{lin2021end-to-end} & EFT~\cite{joo2020eft} & SPIN~\cite{kolotouros2019learning} & SPIN-scene & Ours\\
% \hline \hline
\midrule
MPJPE & 98.5 & 102.1 & 106.5 & 91.6 & \textbf{80.4} \\
PA-MPJPE & 66.9 & 64.8 & 67.1 & \textbf{64.5} & \textbf{64.5} \\
% \hline
 \bottomrule[1pt]
 \end{tabular}
 % \vspace{-0.4cm}
\end{table}

\section{More Qualitative Results}
More qualitative examples and diverse sampling results of our proposed method are shown in Fig.~\ref{fig:appendix-qualitative-examples} and Fig.~\ref{fig:appendix-diversity-samples}, respectively. While obtaining accurate pose estimations aligning with the input image, our method also achieves impressive sample diversity for the unobserved body parts, with plausible human-scene interaction relationships.

%%%%%%%%%%%%
\begin{figure*}[t]
\centering
\includegraphics[width=0.8\textwidth]{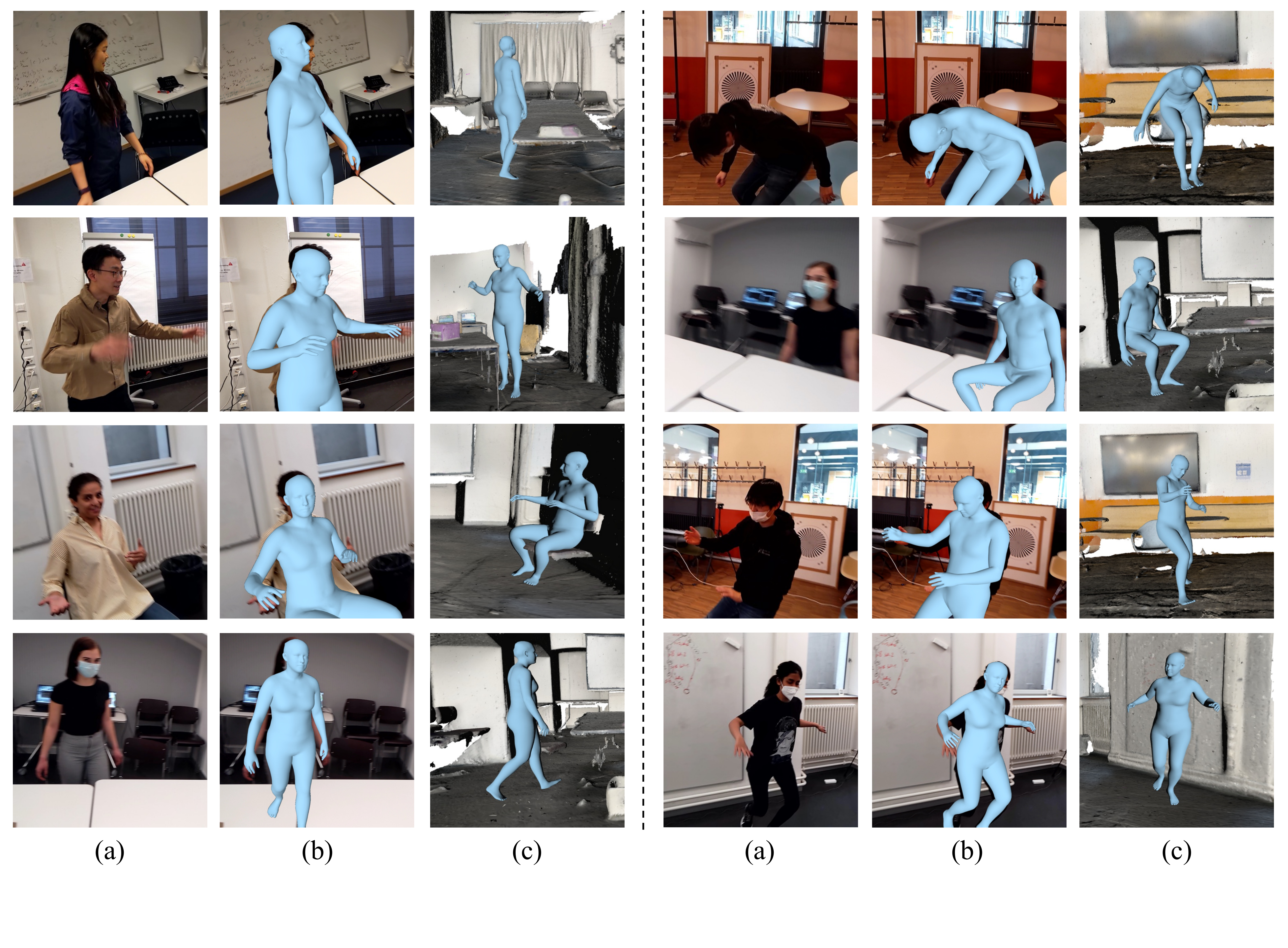}
\vspace{-1cm}
\caption
{
\textbf{More qualitative examples.} For both left and right sides: (a) the input egocentric image; (b) the rendered body mesh overlay on the input image; (c) the rendered body mesh in the 3D scene. 
}
\label{fig:appendix-qualitative-examples}
\end{figure*}
%%%%%%%%%%%%

\vspace{2cm}
\section{Limitations and Future Work}
% temporal
% two stage
Apart from the static human pose, human motions also play an important role in human behavior understanding from the egocentric view. One of the limitations of the proposed method is that it only allows per-frame human mesh recovery given a single frame input. Human motion estimation from an egocentric temporal sequence in 3D scenes would be an exciting future work and enable more real-life AR/VR applications. 
Besides, the current model relies on a two-stage pipeline, which estimates the global body translation and local body pose in separate stages. However, the global translation, local body pose and body shape are coupled together, and equally important for learning the interactions between the human body and the 3D environment. A unified end-to-end model to learn the body parameters altogether would be desired and potentially provide better reasoning about human-scene interaction relationships.

%%%%%%%%%%%%
\begin{figure*}[t]
\centering
\includegraphics[width=\textwidth]{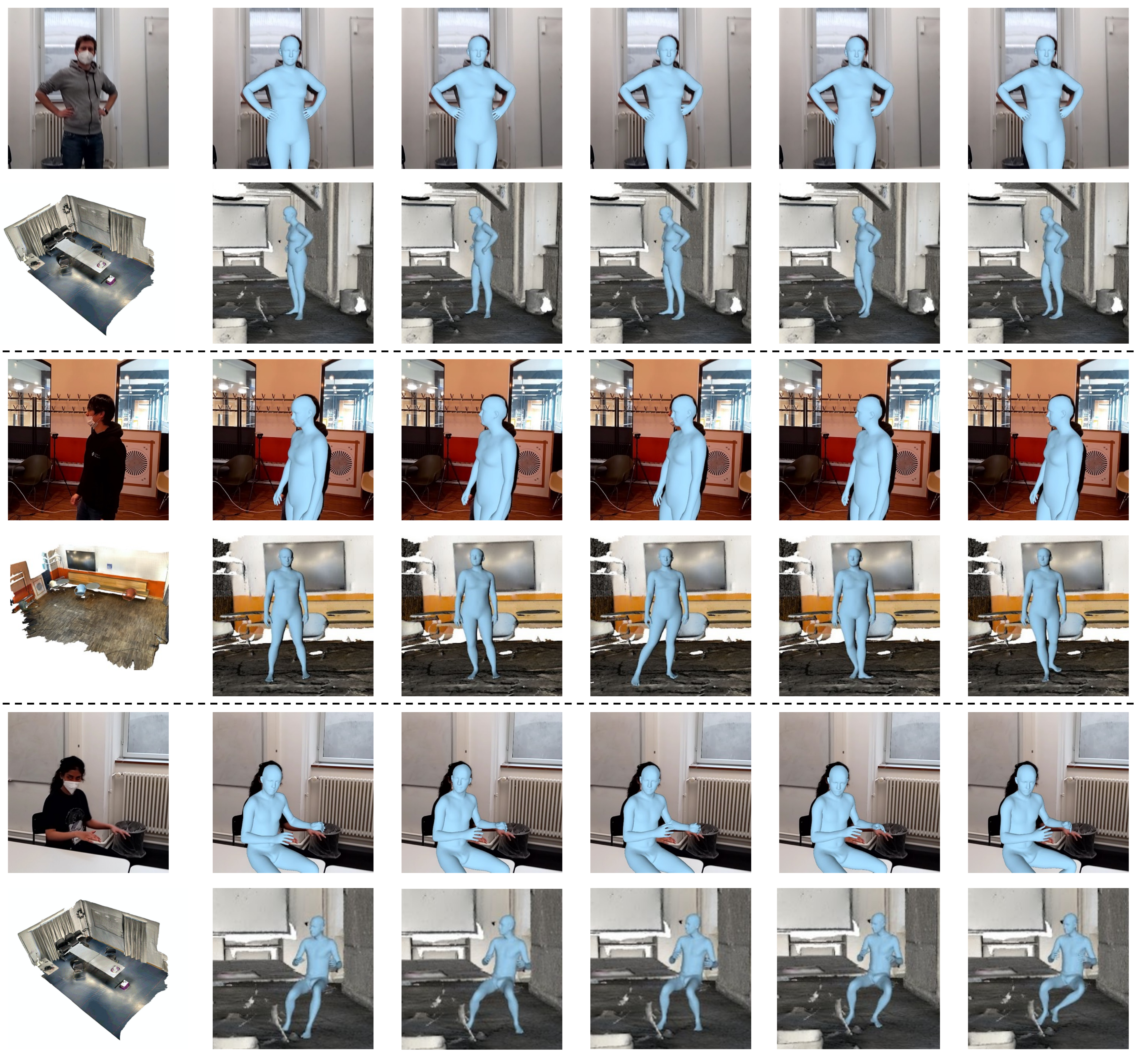}
\caption
{
\textbf{More examples for diverse sampling.} Each row shows five different sample results given the input image and 3D environment (the first column).
}
\label{fig:appendix-diversity-samples}
% \vspace{0.5cm}
\end{figure*}
%%%%%%%%%%%%

\end{document}